# Efficient Inventory Optimization of Multi Product, Multiple Suppliers with Lead Time using PSO

**S.Narmadha**
Assistant Professor
Department of Computer Science and Engineering
Park College of Engineering and Tekhnology
Coimbatore – 641659, Tamilnadu, India

**Dr.V.Selladurai**
Professor and Head
Department of Mechanical Engineering
Coimbatore Institute of Technology
Coimbatore – 641014, Tamilnadu, India

**G.Sathish**
Research Scholar
Department of Computer Science and Engineering
Anna University – Coimbatore, Tamilnadu, India

*Abstract -* With information revolution, increased globalization and competition, supply chain has become longer and more complicated than ever before. These developments bring supply chain management to the forefront of the management's attention. Inventories are very important in a supply chain. The total investment in inventories is enormous, and the management of inventory is crucial to avoid shortages or delivery delays for the customers and serious drain on a company's financial resources. The supply chain cost increases because of the influence of lead times for supplying the stocks as well as the raw materials. Practically, the lead times will not be same through out all the periods. Maintaining abundant stocks in order to avoid the impact of high lead time increases the holding cost. Similarly, maintaining fewer stocks because of ballpark lead time may lead to shortage of stocks. This also happens in the case of lead time involved in supplying raw materials. A better optimization methodology that utilizes the Particle Swarm Optimization algorithm, one of the best optimization algorithms, is proposed to overcome the impasse in maintaining the optimal stock levels in each member of the supply chain. Taking into account the stock levels thus obtained from the proposed methodology, an appropriate stock levels to be maintained in the approaching periods that will minimize the supply chain inventory cost can be arrived at.

## I. INTRODUCTION

Inventory takes many forms, ranging from raw materials to finished goods. While holding large amounts of inventory enables a company to be responsive to fluctuations in customer demand, the associated costs can be excessive. In order to operate in a lean environment at maximum efficiency levels, companies must minimize all unnecessary expenses, including those associated with production and storage of inventories.

Inventory control is typically a key aspect of almost every manufacturing and/or distribution operation business. The ultimate success of these businesses is often dependent on its ability to provide customers with the right goods, at the right place, at the right time. The right goods are those that the customer wants; the right place is your "available" inventory, not the supplier's warehouse, and in today's economy the right time is immediately.

Failure to have the right goods in the right place at the right time often leads to lost sales and profits and, even worse, to lost customers. Today's reality is that there is very little differentiation between commodity products of the same type, and customers will, more often than not, choose to return to businesses that meet all three conditions, even choosing relatively unknown brands over known brands.

The role of inventory management is to coordinate the actions of all business segments, particularly sales, marketing and production, so that the appropriate level of stock is maintained to satisfy customers' demands. The goal of inventory management is to balance supply and demand as closely as possible in order to keep customers satisfied and drive profits.

Inventory management is a fundamental requisite to supply chain optimization. The processes and controls of effective inventory management are critical to any successful business. Since it is rarely the case that any business has the luxury of unlimited capital, inventory management involves important decisions about what to buy or produce, how much to buy or produce and when to buy or produce within the capital limits. These are "value decisions." Excessive inventory investments can tie up capital that may be put to better use within other areas of the business. On the other hand, insufficient inventory investment can lead to inventory shortages and a failure to satisfy customer demand. A balance must be struck and maintained.

The aim of inventory management is to reduce inventory holdings to the lowest point without negatively impacting availability or customer service levels. This can be done while still maximizing the business' ability to exploit economies of scale to positively impact profitability.

Inventory optimization takes inventory management to the next level, enabling businesses to further reduce inventory levels while improving customer service levels and maximizing capital investments.

Inventory management is an ongoing process that relies on inputs from forecasts and product pricing, and should be executable within the cost structure of the business under an overall plan. Inventory control involves three inventory forms of the flow cycle:

- Basic Stock - The exact quantity of an item required to satisfy a demand forecast.

- Seasonal Stock - A quantity buildup in anticipation of predictable increases in demand that occur at certain times in the year.

- Safety Stock - A quantity in addition to basic inventory that serves as a buffer against uncertainty.

The challenge is to weigh the balance in favor of basic stock so that the business holds as little safety stock as possible and provides 'just the right amount' of seasonal stock. However, the predictability of demand has a direct impact on how much safety stock a business must hold. When demand is unpredictable, higher levels of safety stock must be maintained. Therefore, the search for the optimal inventory levels to achieve a lean manufacturing environment becomes a key objective.

*A. Benefits of Inventory Optimization*

The primary function of an Inventory Optimization solution is to allow companies to effectively fulfill demand and identify how to gain additional profits from their inventories. Improved efficiencies through effective resource management and optimization lead to an increase in service level, improved performance against customer request dates and improved return on equity. These gains are derived in three ways: a) System Benefits b) Value-Added Benefits and c) Strategic Benefits.

*B. Particle Swarm Optimization*

In 1995, Kennedy and Eberhartin, inspired by the choreography of a bird flock, first proposed the Particle Swarm Optimization (PSO). In comparison with the evolutionary algorithm, PSO, relatively recently devised population-based stochastic global optimization algorithm, has many similarities and the robust performance of the proposed method over a variety of difficult optimization problems has been proved [1]. In accordance with PSO, either the best local or the best global individual affects the behavior of each individual in order to help it fly through a hyperspace [2]. Simulation of simplified social models has been employed to develop Particle Swarm Optimization techniques. The following are the features of the method [3]:

- The researches on swarms such as fish schooling and bird flocking are the basis of the method.

- The computation time is short and it requires little memory as it is based on a simple concept.

- Nonlinear optimization problems with continuous variables were the initial focus of this method. Nevertheless, problems with discrete variables can be treated by easy expansion of the method. Hence, the mixed integer nonlinear optimization problems with both continuous and discrete variables can be treated with this method.

In addition to PSO, several evolutionary paradigms exist which include Genetic algorithms (GA), Genetic programming (GP), Evolutionary strategies (ES) and Evolutionary programming (EP). Biological evolution is simulated by these approaches which are based on population [4]. Genetic algorithm and PSO are two widely used types of evolutionary computation techniques among the various types of Evolutionary Computing paradigms [5].

PSO and evolutionary computation techniques such as Genetic Algorithms (GA) have many similarities between them. A population of random solutions is used to initialize the system which updates generations to search for optima. Nevertheless, PSO does not have evolution operators such as crossover and mutation that are available in GA.

In PSO, the potential solutions, called particles follow the current optimum particles to fly through the problem space. Every particle represents a candidate solution to the optimization problem. The best position visited by the particle and the position of the best particle in the particle's neighborhood influences its position.

Particles would retain part of their previous state using their memory. The particles would still remember the best positions they ever had even as there are no restrictions for particles to know the positions of other particles in the multidimensional spaces. An initial random velocity and two randomly weighted influences: individuality (the tendency to return to the particle's best previous position), and sociality (the tendency to move towards the neighborhood's best previous position) form each particle's movement [6].

When the neighborhood of a particle is the entire swarm, the global best particle refers to the best position in the neighborhood and in this case, gbest PSO refers the resulting algorithm. Generally, lbest PSO refers the algorithm in cases when smaller neighborhoods are used [5]. A fitness function that is to be optimized evaluates the fitness values of all the particles [6].

PSO uses individual and group experiences to search the optimal solutions. Nevertheless, previous solutions may not provide the solution of the optimization problem. The optimal solution is changed by adjusting certain parameters and putting random variables. The ability of the particles to remember the best position that they have seen is an advantage of PSO [6].

## II. RELATED REVIEW

A fresh Genetic Algorithm (GA) approach for the Integrated Inventory Distribution Problem (IIDP) has been projected by Abdelmaguid et al. [7]. They have developed a genetic representation and have utilized a randomized version of a formerly developed construction heuristic in order to produce the initial random population.

Pongcharoen et al.[8] have put forth an optimization tool that works on basis of a Multi-matrix Real-coded Generic Algorithm (MRGA) and aids in reduction of total costs associated with in supply chain logistics. They have incorporated procedures that ensure feasible solutions such as the chromosome initialization procedure, crossover and mutation operations. They have evaluated the algorithm with the aid of three sizes of benchmarking dataset of logistic chain network that are conventionally faced by most global manufacturing companies.

A technique to utilize in supply-chain management that supports the decision-making process for purchases of direct goods has been projected by Buffett et al.[9]. RFQs have been constructed on basis of the projections for future prices and demand and the quotes that optimize the level of inventory each day besides minimizing the cost have been accepted. The problem was represented as a Markov Decision Process (MDP) that allows for the calculation of the utility of actions to be based on the utilities of substantial future states. The optimal quote requests and accepts at each state in the MDP were determined with the aid of Dynamic programming. A supply chain management agent comprising of predictive, optimizing and adaptive components called the TacTex-06 has been put forth by Pardoe et al. [10]. TacTex-06 functions by making predictions regarding the future of the economy, such as the prices that will be proffered by component suppliers and the degree of customer demand and then strategizing its future actions so as to ensure maximum profit.

Beamon et al.[11] have presented a study on evaluations of the performance measures employed in supply chain models and have also displayed a framework for the beneficial selection of performance measurement systems for manufacturing supply chains Three kinds of performance measures have been recognized as mandatory constituents in any supply chain performance measurement system. New flexibility measures have also been created for the supply chains. The accomplishment of beam-ACO in supply-chain management has been proposed by Caldeira et al.[12]. Beam-ACO has been used to optimize the supplying and logistic agents of a supply chain. A standard ACO algorithm has aided in the optimization of the distributed system. The application of Beam-ACO has enhanced the local and global results of the supply chain.

A beneficial industry case applying Genetic Algorithms (GA) has been proposed by Wang et al.[13]. The case has made use of GAs for the optimization of the total cost of a multiple sourcing supply chain system. The system has been exemplified by a multiple sourcing model with stochastic demand. A mathematical model has been implemented to portray the stochastic inventory with the many to many demand and transportation parameters as well as price uncertainty factors. A genetic algorithm which has been approved by Lo [14] deals with the production-inventory problem with backlog in the real situations, with time-varied demand and imperfect production due to the defects in production disruption with exponential distribution. Besides optimizing the number of production cycles to generate a (R, Q) inventory policy, an aggregative production plan can also be produced to minimize the total inventory cost on the basis of reproduction interval searching in a given time horizon.

Barlas et al.[15] have developed a System Dynamics simulation model of a typical retail supply chain. The intent of their simulation exercise was to build up inventory policies that enhance the retailer's revenue and reduce costs at the same instant. Besides, the research was also intended towards studying the implications of different diversification strategies. A supply chain model functioning under periodic review base-stock inventory system to assist the manufacturing managers at HP to administer material in their supply chains has been introduced by Lee et al.[16]. The inventory levels across supply chain members were obtained with the aid of a search routine.

The inventory and supply chain managers are mainly concerned holding of the excess stock levels and hence the increase in the holding cost. Meanwhile, there is possibility for the shortage of products. For the shortage of each product there will be a shortage cost. Holding excess stock levels as well as the occurrence of shortage for products lead to the increase in the supply chain cost. The factory may manufacture any number of products, each supply chain member may consume a few or all the products and each product is manufactured using a number of raw materials sourced from many suppliers. All these factors pose additional holding of the excess stock levels and hence the increase in the holding cost. Meanwhile, there is possibility for the shortage of products. For the shortage of each product there will be a shortage cost. Holding excess stock levels as well as the occurrence of shortage for products lead to the increase in the supply chain cost. All these factors pose additional challenge in extracting the exact product and the stock levels that influence the supply chain cost heavily.

Many well-known algorithmic advances in optimization have been made, but it turns out that most have not had the expected impact on the decisions for designing and optimizing supply chain related problems. Some optimization techniques are of little use because they are not well suited to solve complex real logistics problems in the short time needed to make decisions and also some techniques are highly problem-dependent which need high expertise. This adds difficulties in the implementations of the decision support systems which contradicts the tendency to fast implementation in a rapidly changing world. IO techniques need to determine a globally optimal placement of inventory, considering its cost at each stage in the supply chain and all the service level targets and replenishment lead times that constraint each inventory location.

## III. OBJECTIVES

The supply chain cost increases because of the influence of lead times for supplying the stocks as well as the raw materials. Practically, the lead times will not be same throughout all the periods. Maintaining abundant stocks in order to avoid the impact of high lead time increases the holding cost. Similarly, maintaining fewer stocks because of ballpark lead time may lead to shortage of stocks. This also happens in the case of lead time involved in supplying raw materials. A better optimization methodology would consider all these above mentioned factors in the prediction of the optimal stock levels to be maintained such that the total supply chain cost can be minimized. Here, an optimization methodology that utilizes the Particle Swarm Optimization (PSO) algorithm, one of the best optimization algorithms, is proposed to overcome the impasse in maintaining the optimal stock levels in each member of the supply chain. Taking into account the stock levels thus obtained from the proposed methodology, an appropriate stock levels to be maintained in the approaching periods that will minimize the supply chain inventory cost can be arrived at.

Supply chain model is broadly divided into four stages in which the optimization is going to be performed. The supply chain model is illustrated in the Fig. 1.

### A. Inventory Optimization of Multiproduct, Multiple Suppliers with Lead Time

Effective supply chain strategies must take into account the interactions at various levels in the supply chain. It is challenging to design and operate a supply chain so that total system wide costs are minimized and system wide service levels are maintained.

Uncertainty is inherent in every supply chain. Supply chains need to be designed to eliminate as much uncertainty as possible to deal effectively with the uncertainty that remains. Demand is not the only source of uncertainty. Delivery lead times, manufacturing yields, transportation times and component/raw material availability can also have significant impact on supply chain. Inventory and back order levels fluctuate considerably across the supply chain, even when customer demand for specific products does not vary greatly. The two desired objectives of improved service and inventory levels seem to be not achieved at the same time since traditional inventory theory tells us that to increase service level, the firm must increase inventory and therefore cost. But recent developments in information and communications technologies, have led to the innovative approaches that allow the firm to improve both the objectives simultaneously.

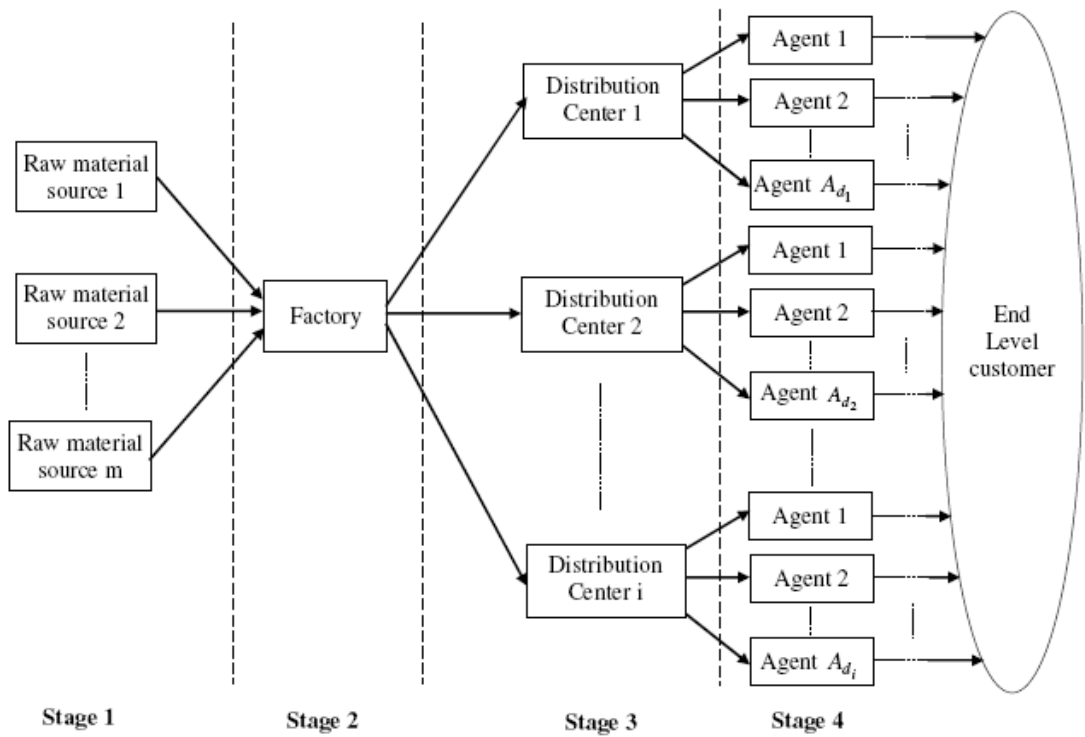

**Fig. 1 Four Stage Supply Chain Model**

In this present research, a prediction analysis that considers all these factors that are mentioned above, such that the analysis paves the way for minimizing the supply chain cost has been proposed. The supply chain cost can be minimized by maintaining optimal stock levels in each supply chain member. Such optimal stock levels can be predicted only by analyzing the past records. This minimization of supply chain cost is effective only if the optimal level is predicted with the knowledge of the lead times of the stocks. Hence a methodology is developed that analyze the past records and predict the emerging excess/shortage of stock levels that are to be considered to identify the optimal stock levels which will be maintained by each of the supply chain member.

Particle Swarm Optimization algorithms, one of the optimization algorithms in Evolutionary computation is used for analysis purpose. The stock levels that are obtained from the analysis are the stock levels that contribute more to the increase of total supply chain cost. These stock levels are used for the prediction of the optimal stock levels that need to be maintained in each supply chain member.

### B. PSO Model for Prediction Analysis

The methodology proposed here will minimize the total supply chain cost by predicting optimal stock levels not only by considering the past records with respect to the stock levels, but also the lead time of the products to reach each supply chain member from its previous stage as well as the lead time taken in supplying the raw materials to the factory. Usually, shortage for a particular stock at a particular member, excess stock levels at a particular member, time required to transport stock from one supply chain member to another i.e. lead time of a stock at a member, time taken to supply raw materials to the factory to manufacture certain products i.e. lead time of raw materials in factory are some of the key factors that play vital role in deciding the supply chain cost. A better optimization methodology should consider all these factors. In the proposed methodology all the above mentioned key factors in predicting the optimal stock levels are considered. Also, different priorities are assigned to those above factors. As per the priority given, the corresponding factors will influence the prediction of optimal stock levels. Hence as per the desired requirement, the optimal stock level will be maintained by setting or changing the priority levels in the optimization procedure.

The optimization is going to be performed in the supply chain model as illustrated in the Fig. 1.

The members participating in the supply chain model are raw material sources $\{r_1, r_2, r_3, \cdots, r_m\}$, a factory $f$, i distribution centers $D = \{d_1, d_2, d_3, \cdots, d_i\}$ and the agents $A = \{A_{d_1}, A_{d_2}, A_{d_3}, \cdots, A_{d_i}\}$, $A_{d_i}$ is the number of agents for the distribution center $d_i$. Hence, the total number of agents in the supply chain model can be arrived using formula :

$$N_A = \sum_{m=1}^{i} A_{d_m} \qquad (1)$$

where $N_A$ is the total number of agents used in the supply chain model.

The factory is manufacturing $k$ number of products. The database holds the information about the stock levels of each product in each of the supply chain member, lead time of products at each supply chain member and lead time of raw material. For $l$ members from factory to end-level-Agents, there are $l-1$ lead times for a particular product and these times are collected from the past records. Similarly, the lead time for raw materials from $r_m$ to $f$ is also taken from the earlier period and thus the database is constituted. Each and every dataset recorded in the database is indexed by a Transportation Identification (TID). For $p$ periods, the TID will be $\{T_1, T_2, T_3, \cdots, T_p\}$. This TID will be used as an index in mining the lead time information.

Now, the particle Swarm Optimization (PSO) is utilized to predict the emerging excess/shortage of stock levels which are vital information for optimal stock levels to be maintained in the future to minimize the supply chain cost. The procedures involved in determining the optimal stock levels are illustrated in Fig. 2.

As the particle swarm optimization (PSO) is more suitable for finding the solution for the optimization problem, PSO is utilized in finding the optimal stock levels to be maintained in each member of the supply chain. The flow of procedures is discussed below.

The individuals of the population including searching points, velocities, $p_{best}$ and $g_{best}$ are initialized randomly but within the lower and upper bounds of the stock levels for all supply chain members, which have to be specified earlier. Hence the generated searching point individual is

$$I_i = [P_k \quad S_1 \quad S_2 \quad S_3 \cdots \cdots S_l], i = 1,2,3\cdots \cdots, N_p \qquad (2)$$

$$\text{where, } P_{L.B} < P_k < P_{U.B}, S_{L.B} < S_l < S_{U.B}$$

$P_{L.B}$, $P_{U.B}$ and $S_{L.B}$, $S_{U.B}$ are the lower and upper bound values of the number of products and stock levels respectively.

The generated population is having the size of $N_p$ i.e. the number of individuals. Since the total number of members that are maintaining the stock levels from $f$ is $l$, the dimension d of each individual is given by

$$d = l + 1 \qquad (3)$$

and hence the equation (2).

Similarly, the initial velocity for the individual will be

$$[v_1 \quad v_2 \quad v_3 \cdots \cdots v_{l+1}], \ v_{min} < v_{l+1} < v_{max} \qquad (4)$$

where $v_{min}$ and $v_{max}$ are the minimum and maximum limit for velocities respectively.

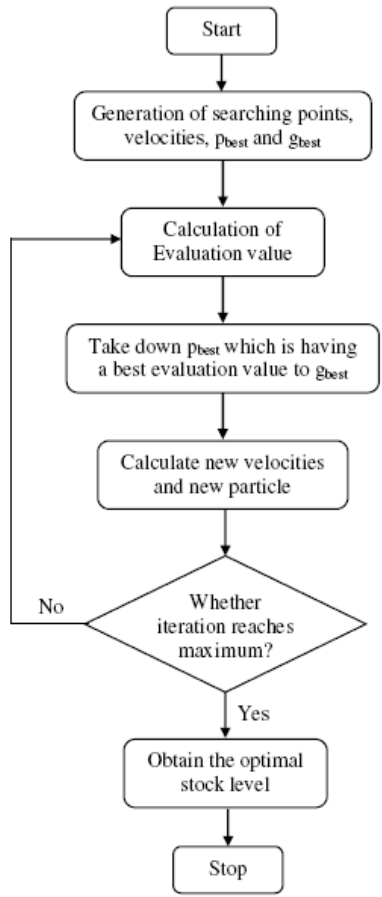

**Fig. 2 Particle Swarm Optimization Steps in Optimizing the Stock Levels**

Then each individual is queried into the database for obtaining the details regarding the TID and frequency of the individual. This will bring $T_q$, $q \in p$ and $P(occ)$, number of periods of occurrence of that particular individual. Here, $q$ need not to be consecutive integers. This obtained TID is queried into the database having the lead time of a particular product to a particular supply chain member. The product I.D. obtained from the individual is queried into the database having the lead time each raw material for the particular product. After all these queries, the lead time of stocks obtained is as follows

$$T_s = [t_{q,1} \quad t_{q,2} \quad \cdots \quad t_{q,l-1}] \tag{5}$$

And the lead time for raw materials is obtained as

$$T_r = [t_1 \quad t_2 \cdots \cdots t_r] \tag{6}$$

where $r$ is the number of raw materials required for a certain product.

Then for each individual the evaluation function is calculated.

**Determination of Evaluation function**

The evaluation function is determined for each randomly generated individual. The function is given by

$$f(a) = w_1 \left(1 - \frac{P(occ)}{T(periods)}\right) + \log(w_2.t_{stock} + w_3.t_{raw})$$

$$a = 1,2,3\cdots\cdots,N_p \tag{7}$$

where $T(periods)$ is the total number of periods of records in database.

In equation (7), $w_1$, $w_2$ and $w_3$ are the weightings of the factors, stock levels, lead time of stocks and lead time of raw materials in optimization, respectively and they are determined as

$$w_1 = \frac{R_1}{R_1 + R_2 + R_3} \tag{8.1}$$

$$w_2 = \frac{R_2}{R_1 + R_2 + R_3} \tag{8.2}$$

$$w_3 = \frac{R_3}{R_1 + R_2 + R_3} \tag{8.3}$$

$R_1$, $R_2$ and $R_3$ are the priority levels of influence of stock levels, lead time of stocks and lead time of raw materials in optimization respectively. Increasing the priority level of a factor increases the influence of the corresponding factor in the evaluation function. Hence this $R_1$, $R_2$ and $R_3$ decides the amount of influence of the factors. The lead time of the stocks $t_{stock}$ is determined as follows

$$t_{stock} = \sum_{i=1}^{l-1} \sum_q t_{q,i} \tag{9}$$

And the lead time required to fill the raw materials is given as

$$t_{raw} = \sum_{i=1}^{r} t_i \tag{10}$$

Equation (9) and (10) is substituted in the equation (7) gives an evaluation value for each individual.

For every individual, a comparison is made between its evaluation value and its $p_{best}$. The $g_{best}$ indicates the most excellent evaluation value among the $p_{best}$. This $g_{best}$ is nothing but an index that points the best individual we have generated so far.

Subsequently the adjustment of the velocity of each particle $a$ is as follows:

$$v_{new}(a,b) = w * v_{cnt}(a) + c_1 * r_1 * [p_{best}(a,b) - I_{cnt}(a,b)]$$
$$+ c_2 * r_2 * [g_{best}(b) - I_{cnt}(a,b)] \tag{11}$$

where,

$$a = 1,2,\cdots\cdots,N_p$$
$$b = 1,2,\cdots\cdots,d$$

Here $v_{cur}(a)$ represents current velocity of the particle, $v_{new}(a,b)$ represents new velocity of a particular parameter of a particle, $r_1$ and $r_2$ are arbitrary numbers in the interval $[0,1]$, $c_1$ and $c_2$ are acceleration constants (often chosen as 2.0), $w$ is the inertia weight that is given as

$$w = w_{max} - \frac{w_{max} - w_{min}}{iter_{max}} \times iter \qquad (12)$$

where $w_{max}$ and $w_{min}$ are the maximum and minimum inertia weight factors respectively that are chosen randomly in the interval $[0,1]$

$iter_{max}$ is the maximum number of iterations

$iter$ is the current number of iteration

Such newly obtained particle should not exceed the limits. This would be checked and corrected before proceeding further as follows,

If $v_{new}(a,b) > v_{max}(b)$, then $v_{new}(a,b) = v_{max}(b)$

if $v_{new}(a,b) < v_{min}(b)$, then $v_{new}(a,b) = v_{min}(b)$

Then as per the newly obtained velocity, the parameters of each particle is changed as follows

$$I_{new}(a,b) = I_{cur}(a,b) + v_{new}(a,b) \qquad (13)$$

Then the parameter of each particle is also verified whether it is beyond the lower bound and upper bound limits. If the parameter is lower than the corresponding lower bound limit then replace the new parameter by the lower bound value. If the parameter is higher than the corresponding upper bound value, then replace the new parameter by the upper bound value. For instance,

If $P_k < P_{L.B}$, then $P_k = P_{L.B.}$

Similarly, if $P_k > P_{U.B}$, then $P_k = P_{U.B.}$

This is to be done for the other parameters also.

This process will be repeated again and again until the maximum number of iterations is reached. Once the maximum number of iterations is attained the process is terminated. The latest $g_{best}$ pointing the individual is the best individual which is having the stock levels that are to be considered and these stock levels are utilized in taking the necessary steps for maintaining the optimal stock levels at each of the supply chain members.

## IV. EXPERIMENTAL RESULTS

The approach suggested for the optimization of inventory level and thereby efficient supply chain management has been implemented in the MATLAB 7.4. The database consists of the records of stock levels held by each member of the supply chain for every period. For implementation purpose, five different products are utilized and these products are in circulation in the seven member supply chain network considered. The sample database which consists of the past records is shown in Table 1.

In the database tabulated in Table 1, the second field comprises of the product Identification (PI) and the other fields are related with the stock levels that were held by the respective seven members of the supply chain network. For example, the second attribute and second field of the database is '3' which refers the Product I.D. '3'. The corresponding fields of the same attribute denote the stock levels of the product I.D. '3' in the respective members of the supply chain. Similarly, different sets of stock levels are held by the database.

**Table 1: Sample data from database of different stock levels**

| TI | PI | F1 | F2 | F3 | F4 | F5 | F6 | F7 |
|----|----|------|------|------|------|------|------|------|
| 1 | 3 | 632 | 424 | 247 | -298 | -115 | 365 | 961 |
| 2 | 5 | -415 | 488 | -912 | 979 | -492 | -922 | 205 |
| 3 | 2 | 369 | -686 | -468 | -807 | 183 | -386 | -228 |
| 4 | 2 | 459 | 289 | -522 | -316 | 130 | -854 | 468 |
| 5 | 3 | -663 | 944 | 856 | 451 | -763 | 657 | 484 |
| 6 | 4 | -768 | -937 | -768 | 242 | 369 | -890 | 289 |
| 7 | 3 | -890 | 619 | -629 | -844 | 791 | 285 | 596 |
| 8 | 3 | 193 | -263 | -474 | 325 | -409 | -216 | -738 |
| 9 | 2 | 578 | -890 | 675 | -321 | -411 | 239 | -916 |
| 10 | 1 | -192 | -421 | 593 | 394 | -141 | 955 | -456 |
| 11 | 5 | 494 | -317 | 600 | 363 | 698 | 927 | 621 |
| 12 | 3 | -838 | -355 | 776 | 682 | -813 | -350 | -513 |
| 13 | 5 | 673 | -593 | 628 | 637 | 643 | 622 | -244 |
| 14 | 1 | -974 | -311 | -319 | -189 | -449 | 663 | 520 |
| 15 | 3 | -725 | 797 | 184 | 406 | -888 | -575 | -144 |
| 16 | 4 | 811 | -569 | -473 | 615 | 467 | -748 | 192 |
| 17 | 1 | -102 | -703 | -859 | 983 | 206 | -803 | -445 |
| 18 | 2 | 426 | 689 | -735 | -465 | 680 | -913 | 147 |
| 19 | 3 | 655 | 626 | -158 | -485 | -622 | -928 | 515 |
| 20 | 5 | -175 | 662 | -847 | 819 | 239 | -902 | 372 |

**Table 2: Sample data from database which is having lead times for stocks**

| TI | T1 | T2 | T3 | T4 | T5 | T6 |
|----|----|----|----|----|----|----|
| 1 | 28 | 27 | 19 | 9 | 19 | 19 |
| 2 | 35 | 33 | 16 | 4 | 24 | 15 |
| 3 | 38 | 38 | 20 | 8 | 10 | 18 |
| 4 | 25 | 25 | 9 | 21 | 22 | 13 |
| 5 | 45 | 40 | 15 | 4 | 16 | 11 |
| 6 | 36 | 43 | 7 | 13 | 21 | 3 |
| 7 | 31 | 27 | 20 | 5 | 22 | 4 |
| 8 | 45 | 28 | 5 | 8 | 10 | 6 |
| 9 | 41 | 45 | 6 | 22 | 8 | 20 |
| 10 | 34 | 26 | 5 | 21 | 12 | 4 |
| 11 | 45 | 38 | 1 | 2 | 11 | 16 |
| 12 | 40 | 38 | 9 | 18 | 22 | 23 |
| 13 | 38 | 40 | 11 | 17 | 7 | 10 |
| 14 | 47 | 47 | 22 | 8 | 23 | 23 |
| 15 | 35 | 27 | 16 | 12 | 22 | 4 |

Sample data from database which is having lead times for stocks is given in Table 2; and sample data set taken from a database which is having raw material lead time for different products is given in Table 3. Initial random individuals for running the PSO algorithm is given in Table 4; and initial random velocities corresponding to each particle of the individual required for the PSO algorithm is given in Table 5.

**Table 3: Raw material lead time for different products**

| PI | RM | T |
|----|----|----|
| 1 | 1 | 20 |
| 1 | 2 | 3 |
| 1 | 3 | 8 |
| 2 | 1 | 10 |
| 2 | 2 | 3 |
| 2 | 3 | 9 |
| 2 | 4 | 23 |
| 2 | 5 | 7 |
| 3 | 1 | 24 |
| 3 | 2 | 23 |
| 3 | 3 | 20 |
| 3 | 4 | 22 |
| 4 | 1 | 3 |
| 4 | 2 | 22 |
| 4 | 3 | 19 |
| 4 | 4 | 18 |
| 5 | 1 | 23 |
| 5 | 2 | 16 |
| 5 | 3 | 23 |
| 5 | 4 | 21 |

**Table 4: Initial Random Individuals**

| PI | F1 | F2 | F3 | F4 | F5 | F6 | F7 |
|----|----|----|----|----|----|----|----|
| 3 | 855 | 61 | 215 | 863 | 24 | 75 | -757 |
| 5 | 854 | -154 | 145 | -241 | -215 | 415 | 845 |

**Table 5: Initial Random Velocities corresponding to each particle of the individual**

| PI | F1 | F2 | F3 | F4 | F5 | F6 | F7 |
|----|----|----|----|----|----|----|----|
| -0.1298 | 0.0376 | -0.3439 | 0.3567 | 0.0982 | -0.0560 | -0.1765 | -0.0409 |
| -0.4997 | 0.0863 | 0.3573 | -0.0113 | 0.0524 | 0.2177 | 0.6550 | 0.0342 |

## V. DISCUSSION OF RESULTS

An iteration involving all these processes was carried out so as to obtain the best individual. Here for instance, the iteration value of '100' is chosen and so hundred numbers of iterative steps will be performed. The best individual obtained as a result is

| 3 | -602 | -280 | -821 | 398 | 382 | -764 | -125 |
|---|------|------|------|-----|-----|------|------|

and its database format is depicted in the table 6.

**Table 6: Database format of Final Best Individual**

| PI | F1 | F2 | F3 | F4 | F5 | F6 | F7 |
|----|----|----|----|----|----|----|----|
| 3 | -602 | -280 | -821 | 398 | 382 | -764 | -125 |

As long as minimization of the fitness function is still possible, then the iteration continues till such a time that no improvement in the evaluation function value is noticeable. After a certain number of iterations, if the evaluation function value is not improving from the previous iterations, then this is an indication that the evaluation function value is stabilizing and the algorithm has converged towards optimal solution. This inference is useful for deciding the number of iterations for running the PSO simulation as well as this may be used as the stopping criteria for the algorithm.

For greater accuracy, the number of iterations should be sufficiently increased and run on the most frequently updated large database of past records. In our experimentation, the iteration value of 100 is chosen and its value is 3.8220 and the values for weighting factors w1, w2 and w3 obtained are 0.6250; 0.3125 and 0.0625 respectively.

The final individual obtained from the PSO based analysis shown in the Table 6 is the inventory level that has potential to cause maximum increase of supply chain cost. It is inferred that controlling this resultant individual is sufficient to reduce the loss either due to the holding of excess stocks or due to the shortage of stocks. By focusing on the excess/shortage inventory levels and initiating appropriate steps to eliminate the same at each member of the chain, the organization can optimize the inventory levels in the upcoming period and thus minimize the supply chain cost.

That is, the organization should focus on the most potential product among multi products that could cause maximum inventory cost and in this specific case it is product 3. The organization should take necessary steps to increase the production of product 3 in the factory by 602 units to make up for the predicted shortage; increase the inventory level of product 3 by 280 units in distribution centre 1 to make up for the predicted shortage; increase the inventory level of product 3 by 821 units in distribution centre 2 to make up for the predicted shortage; decrease inventory level of product 3 by 398 units in agent 1 to make up for the predicted excess; decrease inventory level of product 3 by 382 units in agent 2 to make up for the predicted excess; increase the inventory level of product 3 by 764 units in agent 3 to make up for the predicted shortage; increase the inventory level of product 3 by 125 units in agent 4 to make up for the predicted shortage.

Thus by following the predicted stock levels, the increase in the supply chain cost due to excess/shortage inventory levels can be avoided. The analysis provided an inventory level that had maximum potential to cause a remarkable contribution towards the increase of supply chain cost. The organization can predict the future optimal inventory levels in all the supply chain members with the aid of these levels. Therefore it is concluded that it is possible to minimize the supply chain cost by maintaining the optimal stock levels in the upcoming period among various partners in the supply chain that was predicted from the inventory analysis, making the inventory management further effective and efficient.

## VI. CONCLUSION

Inventory management is an important component of supply chain management. As the lead time plays vital role in the increase of supply chain cost, the complexity of predicting the optimal stock levels increases. The novel and proficient approach based on PSO algorithm, one of the best optimization algorithms, is proposed to overcome the impasse in maintaining the optimal stock levels in each member of the supply chain. The proposed methodology reduced the total supply chain cost as it undoubtedly established the most probable surplus stock level and shortage level along with the consideration of lead time in supplying the stocks as well as raw materials that are required for inventory optimization.

The organizations can make use of the proposed techniques in this present research for inventory optimization by capturing the database in the desired format to suit their respective supply chain structure, replacing the simulated data used in this research with the real data of the organization. For greater accuracy, the number of iterations should be sufficiently increased and run on the most frequently updated large database of past records. Also the organization can mention the maximum possible lower limit and upper limit for the shortage and excess inventory levels respectively, within which the inventory is expected to fluctuate among the various members of the supply chain to make the convergence faster towards optimal solution. Further if the organization can evaluate and quantify the cost involved for each shortage as well as excess of inventory at each member of the supply chain, then the exact savings due to inventory optimization in the supply chain can be calculated for the organization.

## VII. FUTURE SCOPE

Organization can adopt decomposition technique and use the proposed techniques for inventory optimization of high/medium/low value independent products. In this case, the organization should apply the technique for the high value items separately, medium value products separately and low value products separately for inventory optimization in the supply chain.

### ABOUT AUTHORS

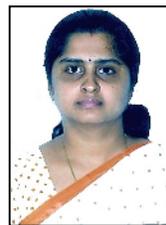

**Mrs. S. Narmadha** is working as Assistant Professor in the Department of Computer Science and Engineering in Park College of Engineering and Technology, Coimbatore. She obtained her Bachelor's degree in Computer Science and Engineering from Tamilnadu College of Engineering, Coimbatore under Bharathiar University and Master's degree in Mechatronics from Vellore Institute of Technology, Vellore. She is currently pursuing Ph.D. under Anna University, Chennai. She has 8 years of Teaching Experience and 2 years of Industrial experience. She has published 14 papers in International Conferences, 2 papers in International journals and a book on 'Open Source Systems'. She is life member of ISTE. Her field of interest includes Supply Chain Management, Automation, Database Management Systems, Virtual Instrumentation, Soft Computing Techniques and Image Processing.

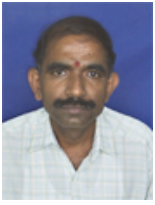

**Dr. V. Selladurai** is the Professor and Head, Department of Mechanical Engineering and Principal, Coimbatore Institute of Technology, Coimbatore, India. He holds a Bachelor's degree in Production Engineering, a Master's degree in Industrial Engineering specialisation and a Ph.D. degree in Mechanical Engineering. He has two years of industrial experience and 22 years of teaching experience. He has published over 90 papers in the proceedings of the leading National and International Conferences. He has published over 35 papers in international journals and 22 papers in national journals. His areas of interest include Operation Research, Artificial Intelligence, Optimization Techniques, Non-Traditional Optimization Techniques, Production Engineering, Industrial and Manufacturing Systems, Industrial Dynamics, System Simulation, CAD/CAM, FMS, CIM, Quality Engineering and Team Engineering.

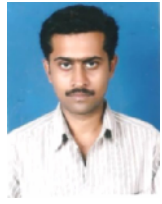

**Mr. G. Sathish** is a full time research scholar under Anna University, Coimbatore. He holds a Bachelor's degree and a Master's degree in Computer Science and Engineering. He has 8 years of industrial experience. He has published 10 papers in the proceedings of the leading International Conferences. His field of interest includes Supply Chain Management, Optimization Techniques, Data Mining, Knowledge Discovery, Automation, Soft Computing Techniques and Image Processing.